\documentclass[preprint,12pt]{elsarticle}

\usepackage[utf8]{inputenc}
\usepackage{fullpage}
\usepackage{natbib}
\usepackage{xcolor}
\usepackage{graphicx}
\usepackage{amsmath}

\usepackage{lineno}
\usepackage{hyperref}
\usepackage{caption}
\usepackage{subcaption}

\journal{arXiv}

\begin{document}
\begin{frontmatter}

\title{The data synergy effects of time-series deep learning models in hydrology}

\author[1,2]{Kuai Fang}
\author[3]{Daniel Kifer}
\author[2]{Kathryn Lawson}
\author[2]{Dapeng Feng}
\author[2]{Chaopeng Shen\corref{cor1}}

\cortext[cor1]{Corresponding author, cshen@engr.psu.edu}

\address[1]{Department of Earth System Science, Stanford University, Stanford, USA}
\address[2]{Department of Civil and Environmental Engineering, Pennsylvania State University, University Park, Pennsylvania, USA.}
\address[3]{Department of Computer Science and Engineering, Pennsylvania State University, University Park, Pennsylvania, USA.}

\begin{abstract}

When fitting statistical models to variables in geoscientific disciplines such as hydrology, it is a customary practice to regionalize - to divide a large spatial domain into multiple regions and study each region separately - instead of fitting a single model on the entire data (also known as unification). Traditional wisdom in these fields suggests that models built for each region separately will have higher performance because of homogeneity within each region. However, by partitioning the training data, each model has access to fewer data points and cannot learn from commonalities between regions. Here, through two hydrologic examples (soil moisture and streamflow), we argue that unification can often significantly outperform regionalization in the era of big data and deep learning (DL). Common DL architectures, even without bespoke customization, can automatically build models that benefit from regional commonality while accurately learning region-specific differences. We highlight an effect we call \emph{data synergy}, where the results of the DL models improved when data were pooled together from characteristically different regions. In fact, the performance of the DL models  \emph{benefited} from more diverse rather than more homogeneous training data. We hypothesize that DL models automatically adjust their internal representations to identify commonalities while also providing sufficient discriminatory information to the model. The results here advocate for pooling together larger datasets, and suggest the academic community should place greater emphasis on data sharing and compilation.

\end{abstract}

\begin{keyword}

\end{keyword}

\end{frontmatter}


\section{Introduction}

As in many other geoscientific fields, there has been a long and pervasive history in hydrology of stratifying data points into different regimes, for which one separately creates statistical models for the variables of interest. This has been done, for example, with hydraulic geometry curves (the relationship between discharge and channel geometries like width and depth): many studies have divided the United States into multiple regions, each of which was fitted with a separate hydraulic geometry curve \cite{Castro2001,Bieger2015}. Regionalized regression formula were used since the early days for annual streamflow \cite{Vogel1999} and evapotranspiration \cite{Fennessey1996}. Regionalization is extensively used for prediction in ungauged basins as a primary means of obtaining parameters for hydrologic models \cite{Razavi2013}. Normally, catchments deemed similar (by spatial proximity, physical similarity, or hydrologic signatures) share hydrologic parameters calibrated from a few of the gauged basins in that group \cite{Beck2016,Guo2021}. In this approach, no information is shared between dissimilar basins. Regionalization is employed to predict flow duration curve \cite{LI2010137}, subsurface stormflow parameters \cite{YE2014670}. For a broader geoscientific example, US is divided into many different fire regimes for modeling wildfires \cite{barrett2010interagency}. The needs for stratification may also partially give popularity to many stratification and classification schemes such as ecoregions, hydrologic landscape regions \cite{Wolock2003} or storage-streamflow response regimes \cite{Fang2017a}.

The well-known learning theory of bias-variance tradeoff \cite{Shalev-Shwartz2014} is at the core of this need for stratification (or regionalization). For a model class (loosely, the set of functions that can be obtained by varying the parameters of a given basic model architecture), bias measures the error of the model that best approximates the underlying true relationship (i.e., the error of the best possible choice of model parameters), while variance measures sensitivity to sampling error and other noise in the training data  (stated another way, model variance measures how much the model parameters can be constrained given the training data at hand). Large variance indicates the model is overfitting to the noise in the data, rather than to the general data trends. Both bias and variance contribute to the overall model error. The bias-variance tradeoff states that if a model class is too simple, it may reduces the variance, but will have a larger bias. On the other hand, if the model class is too complex, it will have a low bias but a large variance, perhaps because there is not enough data to properly constrain the model.

In the framework of the bias-variance tradeoff, the goal of regionalization is to separate out regions with relatively homogeneous conditions so that each region may be characterized by a simple underlying relationship. A small hypothesis class can thus be fitted with acceptable bias. Indeed, Beck et al., 2016 \cite{Beck2016} showed that regionalized parameters performed better than spatially uniform (averaged and calibrated) parameters. In addition, there are always latent variables which cannot be observed or provided as inputs, such as geologic characteristics. Assuming that the important latent variables are relatively homogeneous within each region, their effects can then be conveniently be lumped into the constants and coefficients of the region-specific model. However, if one increases the number of region divisions allowable, the average number of data points per region decreases, thus increasing the variance of each region-specific model. Therefore, one must hope to wisely choose a stratification scheme such that the benefits of simplification due to stratification outweigh the drop in data quantity.

A traditional statistical alternative to this approach is the use of hierarchical and multilevel models. The core concept here is to have both region-specific parameters (to capture commonalities shared within each region) as well as ``global'' parameters that are shared by all regions (to capture commonalities shared between regions). Adapting these ideas to new tasks requires careful consideration of how global and region-specific parameters are likely to interact to form a prediction.

In this paper, we explore an interesting phenomenon associated with deep learning models where a large training set leads to a unified model that is stronger than a collection of regionalized, locally-trained models (i.e., \emph{the whole is greater than the sum of its parts}). We call this effect \emph{data synergy} and hypothesize that deep learning networks use their internal representations to automatically form multilevel models that learn inter-regional homogeneities and heterogeneities (commonalities and differences between regions). This hypothesis has a range of implications. For instance, suppose one is interested in making predictions about region X. One could amass a large homogeneous dataset purely from region X, as well as an equally-sized heterogeneous dataset that contains data not only from X but from other regions as well. According to the theory of data synergy, a model trained on the second dataset should be able to model the commonalities better and should be less prone to overfitting than a model trained on the first dataset. As a result, the data synergy effect would mean that the model trained on this second, heterogeneous dataset to achieve higher predictive performance for region X.
Given the current era of big data, such a phenomenon would suggest that researchers could increasingly benefit from sharing and pooling datasets together, even if the data were to come from outside of an individual researcher's region of interest. 




In this paper, we demonstrate the effect of data synergy with time-series DL models in hydrology for 1) satellite-observed soil moisture and 2) streamflow measured at basin outlets. In these experiments, predictions from local models (trained using data only from inside the respective region), and predictions from global models (trained using more heterogeneous data that include the study region and data from more distant regions), are evaluated in various regions of interest. The experiments are designed to address the following questions: 1) In these applications, are global models better than local models? and 2) Do the models benefit from the diversity of this training data, or simply the increased quantity of training data, or both? The answers to these questions guide us to understand how DL networks work to improve model performance.

We next detail our methods (\S \ref{sec:methods}) and then we present results and discussions (\S \ref{sec:results}) from the soil moisture and streamflow applications.
In \S \ref{sec_further} we consider  limitations  of the data synergy effects.

\section{Methods and data}\label{sec:methods}
In this section, we first present the datasets leveraged in this study (\S \ref{sec_data}), followed by DL model structure (\S \ref{sec_model}) and specific experimental  designs (\S \ref{sec_exp}).
\subsection{Input and target data set}
\label{sec_data}
We investigated the phenomenon of data synergy as applied to two different types of hydrological predictions: soil moisture and streamflow. 
\subsubsection{Soil moisture data}

In the soil moisture experiments, the Soil Moisture Active and Passive (SMAP) satellite mission's Level 3 radiometer product (L3SMP, version 6) was used as the training target. SMAP measures global surface soil moisture ($<$5cm) on a 36 km Equal-Area Scalable Earth Grid (EASE-Grid) based on L-band passive brightness temperature, with a revisit time about every 2-3 days starting in 2015/04/01. Measurement data from from the time period 2015/04/01 to 2016/03/31 was used for training and data from 2016/04/01 to 2018/03/31 was used for testing.

To inform soil moisture prediction, we provide the model with data containing climate forcing time series (meteorological conditions) and geophysical constants. Climate forcing data includes precipitation, temperature, long-wave and short-wave radiation, specific humidity, and wind speed, which are extracted from the North American Land Data Assimilation System phase II (NLDAS-2). Static physiographic data includes land cover classes, surface roughness, and vegetation density extracted from SMAP flags; soil properties like sand, silt and clay percentages, bulk density, and soil water capacity were obtained from the World Soil Information (ISRIC-WISE) database; and normalized difference vegatation indices (NDVI) were obtained from the Global Inventory Monitoring and Modeling System (GIMMS). 

\subsubsection{Streamflow data}
For streamflow experiments, we collected streamflow observations from the U.S. Geological Survey's (USGS) National Water Information System (NWIS) database. Here our goal was to predict basin runoff (mm/year), which we calculated by dividing USGS streamflow observations recorded at the basin outlet by the area of the basin. The training period was 1979/01/01 to 2009/12/31, and the testing period was 2010/01/01 to 2019/12/31. We selected 2773 USGS basins for which observations were available for more than 90\% of the days in both training and testing periods. Among those basins, 576 of them were categorized as reference basins, which are considered to have low human impacts and high data quality. We re-assembled this dataset, instead of relying on existing datasets such as Catchment Attributes and Meteorology for Large-Sample Studies (CAMELS) \cite{Newman2015}, so that our experiments could use more basins than the 671 basins in CAMELS.

As with the soil moisture dataset, we extracted basin-averaged climate forcings and geophysical attributes as input predictors. For streamflow, however, the daily climate forcing data were extracted from the gridMET \cite{Abatzoglou2013} product, which contains precipitation, temperature, humidity, radiation, and reference evaportranspiration, with a spatial resolution of 1~/24 degree. For each targeted USGS site, we integrated the gridMET dataset with the drainage basin boundary from the Geospatial Attributes of Gages for Evaluating Streamflow II dataset (GAGES-II) \cite{Falcone2011} database. Geographic attributes were also extracted from GAGES-II, and we selected 17 fields likely to impact the rainfall-runoff process, including drainage area, basin compactness ratio, snow percent of precipitation, stream density, percent of first-order stream, base flow index, subsurface flow contact time, dam density, permeability, water table depth, rock depth, slope, as well as dominant ecoregion, nutrient region, geology region, hydrologic landscape, and land cover.

\subsection{Model architecture}
\label{sec_model}

Long-short term memory (LSTM) networks \cite{Hochreiter1997} are general-purpose DL time series models that have also proved to be effective in hydrology applications \cite{Fang2017, Fang2019a, Feng2020, Xiang2020, Kratzert2019}. In this study, we use it to predict two dynamical hydrologic variables (soil moisture and streamflow) using the inputs described in \S \ref{sec_data}. LSTM models were trained on pixels for soil moisture data and on basins for streamflow data. In both cases, we used the same network architecture, which consisted of a linear layer of 256 nodes with rectified linear unit (ReLU) activation, followed by a LSTM layer with 256 nodes, and then a linear output layer. The loss function, or metric which was the models' primary goal to minimize, was root-mean-square error (RMSE) between observed and predicted values. The network was trained to minimize RMSE using the AdaDelta optimizer \cite{Zeiler2012} which dynamically tunes learning rate through training iterations. For soil moisture models, the length of time step was set to be 30 (days) and batch size was 100; for streamflow models, time step length was 365 (days) and batch size was 500. All models were trained for 500 epochs. 

%

\subsection{Experimental design}
\label{sec_exp}
\begin{figure}[h]
\includegraphics[width=1\linewidth]{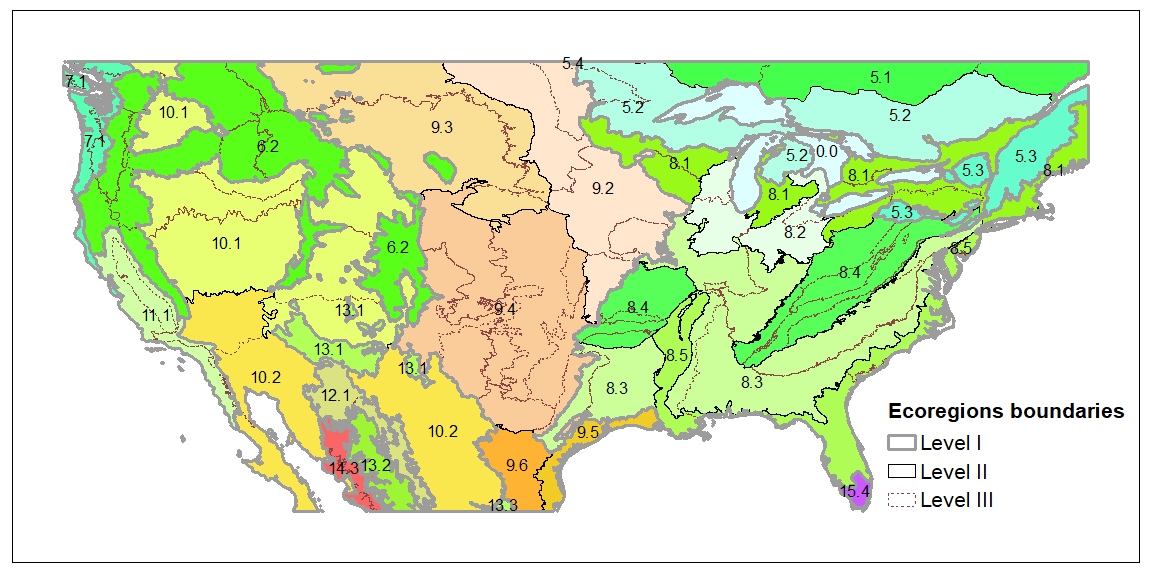}
\includegraphics[width=1\linewidth]{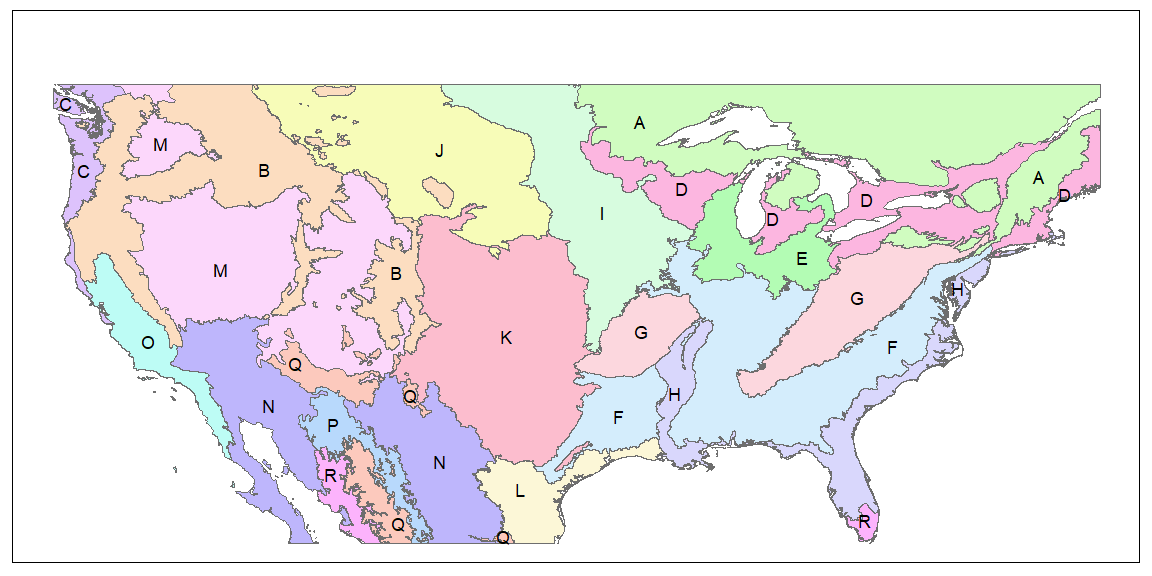}
\caption{a) Map of EPA ecoregions colored based on level II regions. b) Map of the 18 sub-regions based on EPA ecoregions.}
\label{fig_map}
\end{figure}

Stratification of the data was guided by the United States Environmental Protection Agency (EPA) ecoregions \cite{}, as these groupings were devised to provide similarity in terms of surface hydrologic responses. The conterminous United States (CONUS) was divided into ecoregions based on the compositions of geology, landforms, soils, vegetation, climate, land use, wildlife, and hydrology \cite{Omernik2014}. Three hierarchical levels (denoted as I, II and III) of ecoregions divide the CONUS into 11, 25, and 105 regions, respectively. For example, ecoregion 8.3.5 (Southeastern Plains) is a level III ecoregion nested within ecoregion 8.3 (Southeastern USA Plains), which is a level II ecoregion inside level I ecoregion 8 (Eastern Temperate Forests). Figure \ref{fig_map} shows a map of level II EPA ecoregions and the boundaries of ecoregions from level I to III.

The first set of experiments, which we refer to as ``global vs. local'' experiments, compare regionalization (stratifying the data and separately building models on each stratum) to unification (training a single model on the entire data). However, if in these experiments the global model was found to perform better than each of the local models, it could be argued that this was simply because the global model had more data to work with. Thus, the second set of experiments, which we refer to as ``similar vs. dissimilar'' experiments, were designed to test whether the \emph{quantity} of data fully explained the differences between the models, or if the \emph{diversity} of the dataset was also important. For both sets of experiments, the resulting models were evaluated inside various regions of interest (ROI) using temporal generalization tests, where the testing data came from a different time period than the training data (see \S \ref{sec_data}).

\subsubsection{Global vs. local experiments.}\label{mtd_exp_glo}
These experiments were devised to directly compare unification and regionalization.
In order to create regions of approximately equally sizes, we merged ecoregions 14 and 15 since both of them were tropical forests and ecoregion 15 was too small, covering only 0.1\% of the CONUS. We also split apart ecoregions 8, 9, and 10 based on their level II subdivisions, as these three ecoregions were much larger than the others (covering around 70\% of the CONUS). The resulting 18 ``sub-regions'', referred to using letters A-R, had more similar areas between $1\times10^5$ km$^2$ and $1\times10^6$ km$^2$, with an average area of $5\times10^5$ km$^2$ (Figure \ref{fig_map}, Table \ref{tab_ecoid}).

Regions L, N, P, and R, were excluded from the streamflow analyses, as there were almost no reference basins present in those regions. 

\begin{table}[h]
\caption{Conversion between the experimental sub-regions and EPA ecoregions}
\label{tab_ecoid}
\begin{tabular}{ll|ll|ll}
new ID & EPA ID & new ID & EPA ID   & new ID & EPA ID    \\ \hline
A      & 5      & G      & 8.4      & M      & 10.1      \\
B      & 6      & H      & 8.5      & N      & 10.2      \\
C      & 7      & I      & 9.2      & O      & 11.1      \\
D      & 8.1    & J      & 9.3      & P      & 12.1      \\
E      & 8.2    & K      & 9.4      & Q      & 13        \\
F      & 8.3    & L      & 9.5, 9.6 & R      & 14, 15
\end{tabular}
\end{table}

We then compared two scenarios: (1) single LSTM model trained with data from all 18 sub-regions (1 global model), and (2) individual models for each sub-region trained only with data from that sub-region (18 local models). In the testing phase, for each sub-region we compared the predictions from the global model and from that sub-region's corresponding local model. More specifically, the global model was tested on the same pixels (for soil moisture) or gages (for streamflow) inside each sub-region as the corresponding local model. 

\subsubsection{Similar vs. dissimilar experiments}\label{mtd_exp_sim}
The second set of experiments was designed to study the effect of training data composition on model accuracy. Put more simply, if we are interested in creating a prediction model for a ROI, should we gather additional data from nearby/similar regions, or should we instead obtain a more diverse dataset? We used the hierarchical nature of the EPA ecoregions as a proxy for (dis)similarity: two level III ecoregions were defined as being close neighbors if they belonged to the same level II ecoregion, far neighbors if they belonged to the same level I ecoregion (but different level II ecoregions), or dissimilar if they belong to different level I ecoregions. 

For soil moisture, the location of a gridcell centroid determined its ecoregion membership. For streamflow, we determined ecoregion membership based on which ecoregion covered the majority of the basin. Once the streamflow and soil moisture data was categorized by its respective ecoregion, it was clear that the amount of data available for each level III ecoregion varied by a significant amount, and not all of them contained enough data to create viable local models. Thus we selected a subset of level III ecoregions to serve as our regions of interest (ROIs): for soil moisture we selected the six largest level III ecoregions (8.3.5, 9.3.3, 9.4.1, 9.4.2, 10.1.5, and 10.2.4), and for streamflow we selected twelve level III ecoregions containing at least 60 USGS basins (5.3.1, 8.1.7, 8.2.3, 8.2.4, 8.3.1, 8.3.4, 8.3.5, 8.4.1, 8.4.2, 8.5.3, 9.2.3, 9.4.2). 

We compared two scenarios where (1) data size was not controlled (hence the sizes of the datasets were only limited by availability of data), and (2) data size was controlled (so that the homogeneous training data and the heterogeneous data were of roughly the same size). These scenarios are representative of data collection in practice, where there is often a fixed budget for collecting training data, or data outside the region of interest is much more plentiful.

For each ROI (e.g., ecoregion 8.3.5 without data size controlled), we trained four models:
\begin{enumerate}[(1)]
    \item ``local'' model trained on data only from within the ROI (e.g., data from ecoregion 8.3.5).
    \item ``local + close neighbors'' model trained on data from all close neighbors of the ROI, equivalent to the entire level II ecoregion containing the ROI (e.g., data from ecoregion 8.3).
    \item ``local + far neighbors'' model trained using all the far neighbors of the ROI but not the close neighbors (e.g., data from ecoregions 8.3.5, 8.1, 8.2, 8.4, 8.5).
    \item ``local + dissimilar'' model trained using all of the ecoregions that were dissimilar from the ROI (e.g., data from ecoregion 8.3.5 and all areas \emph{outside} of ecoregion 8).
 \end{enumerate}
In this first scenario where training size was not controlled, the models were trained using all of the data in the ecoregions that were available to them.

As mentioned earlier, to help disentangle the impacts of ``more data" and ``more dissimilar data", we trained an additional four models for each ROI where the amount of added training data was controlled. Here, the data points fulfilling the critera for addition beyond the "local" scenario were randomly sampled so that the ``local + close neighbors" , ``local + far neighbors", and ``local + dissimilar" datasets each had the same amount of added data. This modification was performed for the soil moisture data, as the pixels are approximately evenly and continuously spatially distributed, making it straightforward to uniformly sub-sample data from the close, far, and dissimilar regions. This was not the case for streamflow, and consequently these second set of experiments with data size controlled could not be run for streamflow.

\subsubsection{Evaluation of model}
Trained models were evaluated by temporal generation test inside each ROI, on  identical pixels or basins. Soil moisture models were trained from 2015/04/01 to 2017/04/01 and tested from 2017/04/01 to 2018/04/01; streamflow models were trained from 1979/01/01 to 2009/01/01 and tested from 2010/01/01 to 2019/01/01. To evaluate soil moisture models, we calculated the correlation coefficient and root-mean-square error (RMSE) between the observations and predictions for each pixel in a region during the testing period. For streamflow predictions, correlation was also calculated, but the Nash–Sutcliffe model efficiency coefficient (NSE, which is one minus the ratio of model squared error to time series variance) was calculated instead of RMSE, to be in line with previous hydrologic literature. In both cases, the larger the value of the metric, the better a model performs. 

\section{Results and Discussion}\label{sec:results}

\subsection{Global vs. local experiments}
\label{sec_global}
\begin{figure}[h]
\includegraphics[width=1\linewidth]{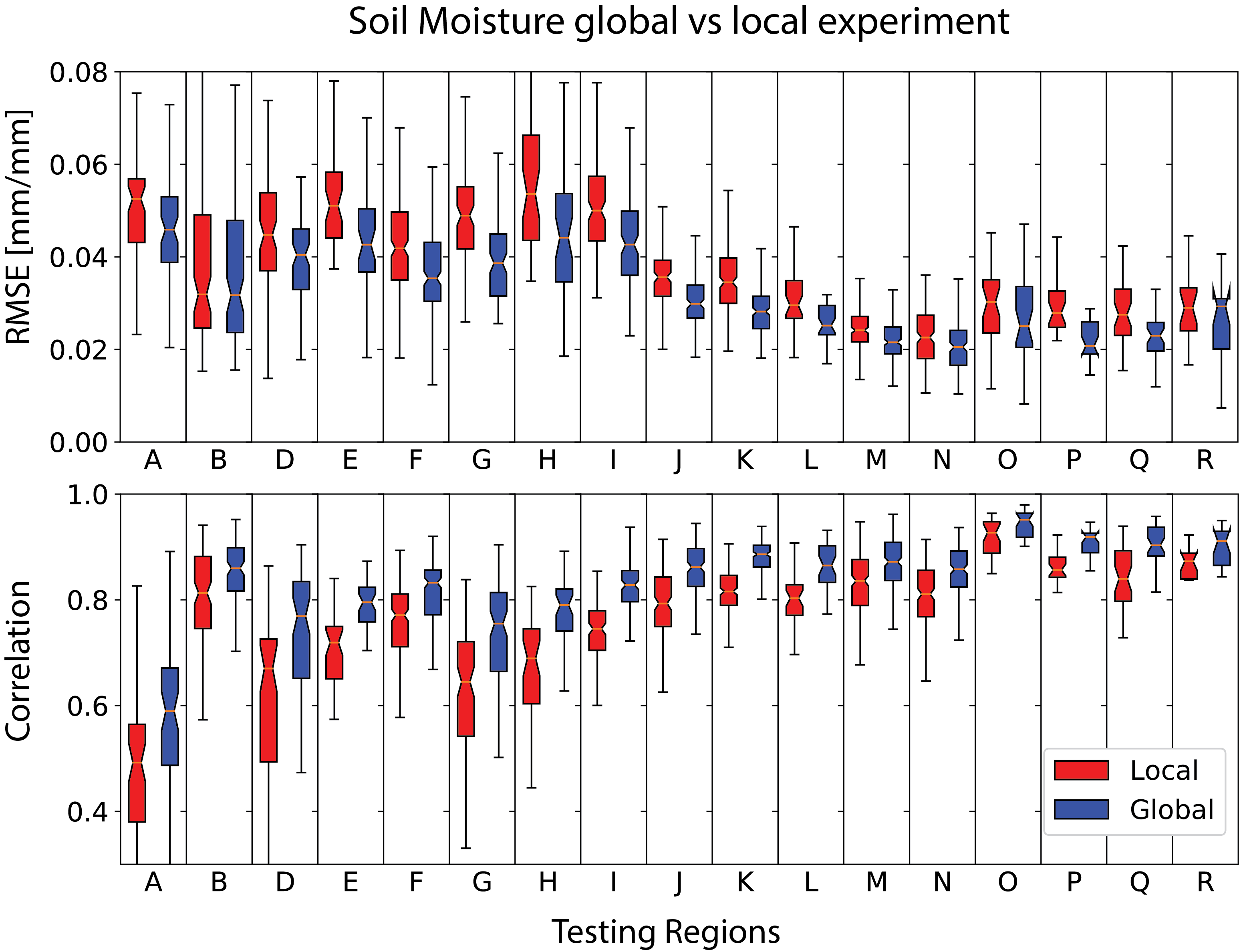}
\caption{Performance metrics of the soil moisture global vs local experiment Upper panel: RMSE; lower panel: correlation. }
\label{fig_global_sm}
\end{figure}

\begin{figure}[h]
\includegraphics[width=1\linewidth]{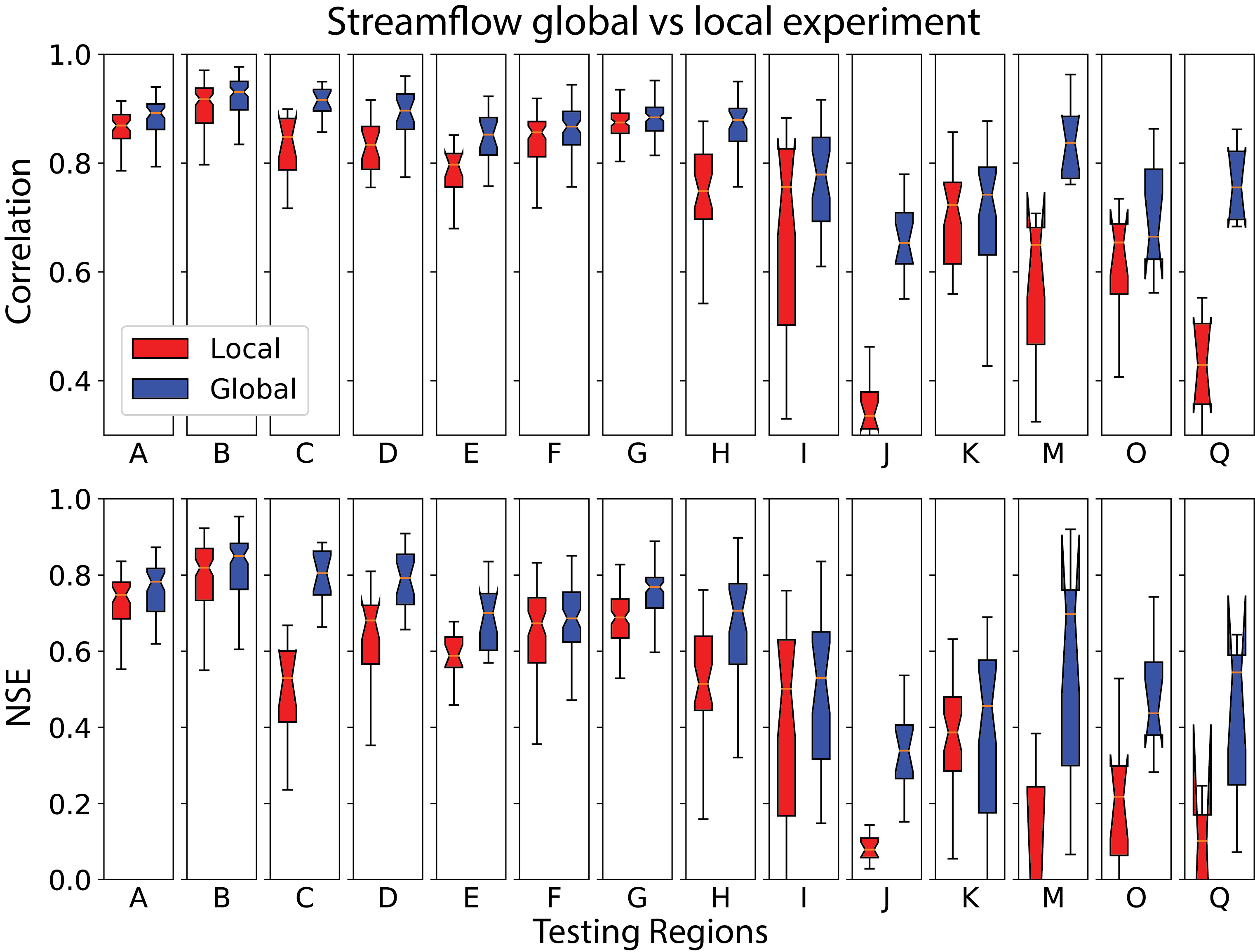}
\caption{Performance metrics of the streamflow global vs local experiment. Upper panel: correlation; lower panel: NSE.}
\label{fig_global_streamflow}
\end{figure}

The global vs. local experiments compared unification (training a single model on the entire dataset) to regionalization (stratifying data by region and separately building models for each individual region). Metrics resulting from these experiments are plotted in Figure \ref{fig_global_sm} for soil moisture and Figure \ref{fig_global_streamflow} for streamflow. Please note that not all regions had pixels (for soil moisture) or basins (for streamflow) present in sufficient number for analysis, so the specific regions investigated will differ (see (\S \ref{mtd_exp_glo}) for details).

For the soil moisture problem, the median RMSE was smaller for the global model than the local model, while the median correlation was larger for each region (Figure \ref{fig_global_sm}). Since we could not assume normality of the metrics, we used the Wilcoxon signed-rank test to measure the statistical significance for each region individually and all together (i.e., when the test is run on data pooled from all regions). These results (p-value and testing sample size) are shown in Table \ref{tab_global}. All of the p-values were small; the largest value was under 0.009 and most were orders of magnitude smaller. Aggregating all the tested pixels, the average test RMSE values for the global and local models were respectively 0.32 and 0.38, while corresponding correlations were 0.82 and 0.75. Global model prediction had a smaller testing RMSE than the local model for 87\% of pixels, and higher correlation for 95\% of pixels. This clearly demonstrates that for soil moisture, the global model consistently and significantly (both in a practical and statistical sense) outperformed the local models.

For the streamflow problem, within each region, the median NSE value (calculated over all basins in the region) for the global model was also larger than that for the local model (Figure \ref{fig_global_streamflow}). It should be noted however that in region K, even though the median NSE was higher, the global model's error variability was so large that in practice the local model would be preferred. As with soil moisture, we used the Wilcoxon signed-rank test to measure the statistical significance (Table \ref{tab_global}). Only regions K and Q had p-values larger than 0.01 (note that region Q only had a sample size of 7 basins). The overall median correlations for the global and local models were 0.84 and 0.79 respectively, while the corresponding NSE values were 0.73 and 0.65. NSE for the global model was higher than for the local model in 81\% of the basins and correlation was higher for 84\% of the basins. Like the soil moisture models, these streamflow modeling results showed that the global model generally had higher quality than the regionalized models.

\begin{table}[]
\caption{Table of p-value in Wilcoxon signed-rank test for global vs local experiments}
\begin{tabular}{|l||l|l|l||l|l|l|}
\hline
& \multicolumn{3}{c||}{Soil Moisture} & \multicolumn{3}{c|}{Streamflow} \\ \hline
\multicolumn{1}{|c||}{Region} 
& \multicolumn{1}{c|}{p for RMSE} & \multicolumn{1}{c|}{p for Corr} & \multicolumn{1}{c||}{N} & \multicolumn{1}{c|}{p for Corr} & \multicolumn{1}{c|}{p for NSE} & \multicolumn{1}{c|}{N} \\ \hline
A & 3.87E-04 & 1.68E-10 & 65 & 1.00E-09 & 8.33E-08 & 54 \\ \hline
B & 6.74E-03 & 1.19E-16 & 105 & 9.78E-14 & 3.55E-12 & 96 \\ \hline
C & No Pixels & No Pixels & 0 & 6.55E-04 & 8.05E-04 & 15 \\ \hline
D & 7.27E-10 & 2.43E-13 & 71 & 1.82E-05 & 3.18E-04 & 24 \\ \hline
E & 1.83E-08 & 6.76E-08 & 43 & 1.82E-04 & 2.90E-03 & 19 \\ \hline
F & 2.21E-29 & 8.39E-32 & 192 & 8.84E-07 & 3.78E-03 & 74 \\ \hline
G & 6.64E-18 & 6.24E-18 & 99 & 8.29E-06 & 5.99E-11 & 89 \\ \hline
H & 3.34E-08 & 7.76E-09 & 45 & 1.68E-07 & 1.31E-06 & 36 \\ \hline
I & 1.26E-19 & 2.41E-21 & 120 & 1.26E-04 & 4.75E-03 & 32 \\ \hline
J & 4.91E-24 & 4.60E-24 & 136 & 4.38E-04 & 7.17E-03 & 16 \\ \hline
K & 2.18E-33 & 2.02E-33 & 193 & 9.03E-02 & 9.57E-01 & 40 \\ \hline
L & 1.94E-06 & 1.34E-07 & 37 & No Basins & No Basins & 0 \\ \hline
M & 4.05E-17 & 9.00E-27 & 194 & 3.70E-03 & 2.22E-03 & 12 \\ \hline
N & 3.75E-17 & 1.72E-23 & 158 & No Basins & No Basins & 0 \\ \hline
O & 4.12E-05 & 5.91E-06 & 33 & 7.65E-03 & 4.44E-03 & 11 \\ \hline
P & 3.79E-06 & 3.79E-06 & 28 & No Basins & No Basins & 0 \\ \hline
Q & 7.27E-07 & 6.14E-08 & 39 & 1.80E-02 & 1.80E-02 & 7 \\ \hline
R & 8.90E-03 & 2.50E-04 & 19 & No Basins & No Basins & 0 \\ \hline
All & 1.08E-201 & 2.39E-244 & 1577 & 4.34E-58 & 3.18E-45 & 525 \\ \hline
\end{tabular}
\label{tab_global}
\end{table}

\subsection{Similar vs. dissimilar experiments}
\label{sec_similar}


\subsubsection{Dataset size not controlled}
\label{sec_similar_no_control}
In the ``similar" vs. ``dissimilar" experiments, any ecoregion will clearly have more ``far'' neighbors than ``close'' neighbors, and more ``dissimilar'' regions than ``far'' and ``close'' neighbors. When the training set size was not controlled, the question tested was ``what helps more when added to the training set, additional data that are as similar as possible to the ROI (indicating a preference for homogeneity), or a larger, more diverse dataset?" This is of particular interest in situations when heterogeneous data are plentiful and easier to collect than homogeneous data. 

\begin{figure}[h]
\includegraphics[width=1\linewidth]{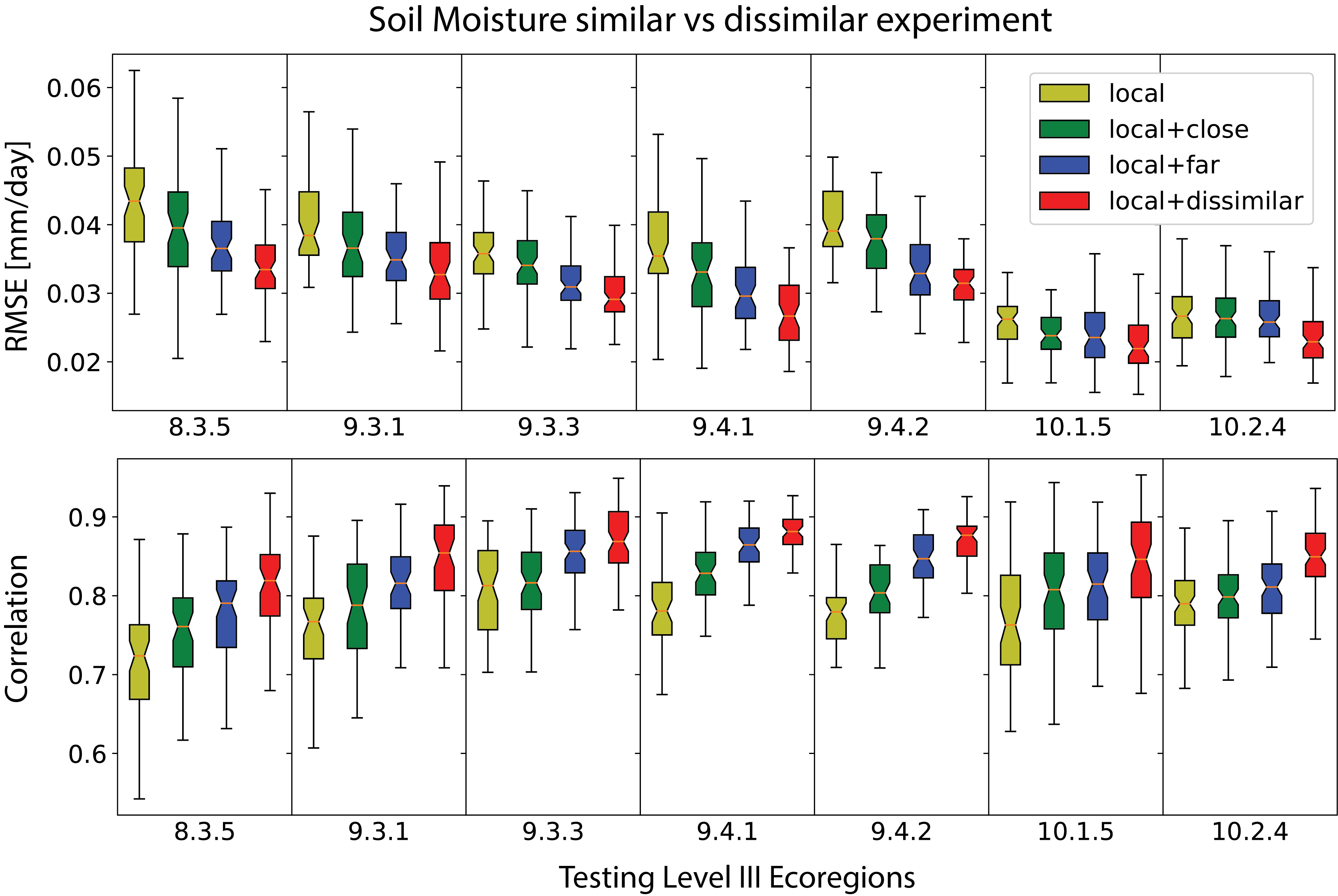}
\caption{Performance metrics of the soil moisture similar vs dissimilar experiment without controlling for data size. Upper panel RMSE; lower panel: correlation. }
\label{fig_sim_sm}
\end{figure}

For SMAP soil moisture prediction in each of the chosen ROI, we saw that RMSE and correlation monotonically improved as we added increasingly diverse data to the "baseline" local model (the model trained using only data from the ROI), with the best performance being achieved by the most heterogeneous dataset (local + dissimilar) (Figure \ref{fig_sim_sm}). The improvement was less pronounced for the drier western regions (10.1.5, 10.2.4) as compared to the wetter eastern regions, where soil moisture has larger fluctuations. After evaluating statistical significance, we saw that in these wetter regions, all of the pairwise comparisons were significant, with p-values much lower than the 0.01 significance threshold (Table \ref{tab_sim_sm}). For the two drier regions (10.1.5, 10.2.4), a few of the comparisons were not statistically significant at this small sample size. However, for correlation, all comparisons involving ``local + dissimilar'' were significant, showing that adding in the data from other Level I ecoregions not only did not hurt performance (as conventional wisdown might suggest), but actually helped the most.

\begin{figure}[h]
\includegraphics[width=1\linewidth]{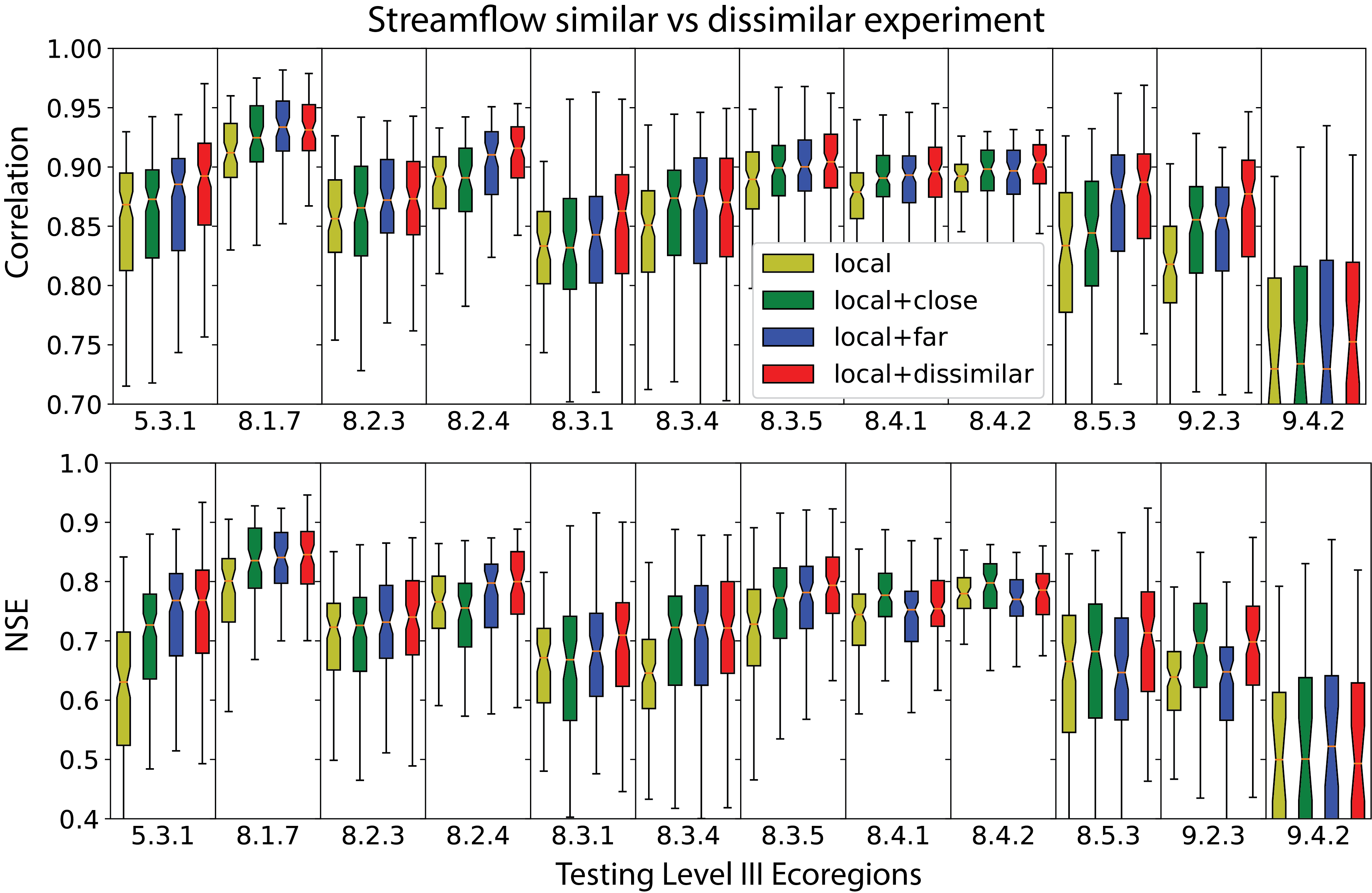}
\caption{Performance metrics of the streamflow similar vs dissimilar experiment without controllling for dataset size. Upper panel correlation; lower panel: NSE.}
\label{fig_sim_streamflow}
\end{figure}

For streamflow we observed a similar general trend in that a more diversified training set improved predictions, but the effect was smaller than for soil moisture and not as monotonic (Figure \ref{fig_sim_streamflow}). Due to the smaller effect size and the small sample size within each region, most, but not all comparisons were statistically significant at the 0.01 level (Table \ref{tab_sim_q}). However, when the ROI were pooled together for hypothesis testing (last line of Panels A and B), they showed unambiguously that the differences were statistically significant, implying that overall, diversity helped improve predictions.

There were some exceptions to this trend, however. Upon closer inspection, we noted that for some cases, NSE dropped from ``local + close'' to ``local + far'' data (regions 8.4.1, 8.4.2, 9.2.3), suggesting that in those cases, the dissimilar training set may have introduced additional bias to the model (Figure \ref{fig_sim_streamflow}). Furthermore, when the LSTM models performed poorly (e.g., region 9.4.2), including diverse training regions did not improve the model performance. Large errors tended to be associated with large basin areas, which may have been due to a variety of factors including (1) the subbasins were heterogeneous and there was not enough data for the local model to learn this heterogeneity, (2) the watershed boundaries were unclear, or (3) cross-basin groundwater flow (which was not part of the model) could have a larger impact than anticipated.

These observations suggest that one needs to prioritize the collection of enough local data to built a local model with reasonably good performance. After that, additional improvements can be obtained from data collected outside the ROI with preference towards heterogeneous data, as it may provide a regularizing effect and help guard against overfitting. If the local model \emph{under}fits though, the heterogeneous data may not help. It is worth repeating, however, that while this was the case for 'close' vs. 'far' regions, comparisons between 'close' vs. 'dissimilar' always showed significantly improved predictions.

\begin{table}[]
\caption{P-value in Wilcoxon signed-rank test for similar vs dissimilar soil moisture experiments without controlling for data size. For space reasons, ``local + close" was abbreviated as ``close'' and similarly for ``local+far'' and ``local+dissimilar''}
\subcaption*{Panel A: p-value for RMSE}
\begin{tabular}{|c|c|c|c|c|c|}
\hline
Ecoregion III & local vs close & close vs far & far vs dissimilar & close vs dissimilar & N \\ \hline
8.3.5 & 9.35E-09 & 9.44E-08 & 5.29E-09 & 1.63E-11 & 60 \\ \hline
9.3.1 & 9.00E-08 & 3.40E-07 & 2.39E-06 & 6.64E-10 & 52 \\ \hline
9.3.3 & 2.40E-04 & 1.55E-12 & 1.37E-08 & 7.64E-13 & 68 \\ \hline
9.4.1 & 2.66E-10 & 5.95E-10 & 3.49E-09 & 1.31E-10 & 55 \\ \hline
9.4.2 & 3.07E-07 & 3.30E-10 & 9.51E-07 & 1.11E-10 & 55 \\ \hline
10.1.5 & 1.00E-09 & 6.53E-01 & 4.51E-07 & 6.09E-07 & 61 \\ \hline
10.2.4 & 8.11E-03 & 9.63E-06 & 5.14E-14 & 1.89E-14 & 78 \\ \hline
All & 2.44E-44 & 1.09E-42 & 9.94E-52 & 4.56E-67 & 429 \\ \hline
\end{tabular}
\subcaption*{Panel B: p-value for correlation}
\begin{tabular}{|c|c|c|c|c|c|}
\hline
Ecoregion III & local vs close & close vs far & far vs dissimilar & close vs dissimilar & N \\ \hline
8.3.5 & 3.55E-09 & 8.44E-07 & 9.35E-09 & 2.10E-11 & 60 \\ \hline
9.3.1 & 5.71E-08 & 2.28E-06 & 3.28E-05 & 1.85E-09 & 52 \\ \hline
9.3.3 & 5.10E-05 & 2.97E-11 & 1.52E-06 & 3.41E-12 & 68 \\ \hline
9.4.1 & 2.32E-09 & 4.72E-09 & 1.55E-08 & 1.63E-10 & 55 \\ \hline
9.4.2 & 3.83E-07 & 8.62E-10 & 1.20E-04 & 1.11E-10 & 55 \\ \hline
10.1.5 & 4.17E-05 & 1.95E-01 & 1.87E-01 & 1.94E-02 & 61 \\ \hline
10.2.4 & 5.65E-05 & 8.75E-01 & 4.06E-11 & 1.93E-11 & 78 \\ \hline
All & 7.05E-41 & 7.08E-34 & 1.36E-35 & 5.38E-58 & 429 \\ \hline
\end{tabular}
\label{tab_sim_sm} 
\end{table}

\begin{table}[]
\caption{P-value in wilcoxon signed-rank test for similar vs dissimilar streamflow experiments without controlling for data size. For space reasons, ``local + close" was abbreviated as ``close'' and similarly for ``local+far'' and ``local+dissimilar''}
\subcaption*{Panel A: p-value for correlation}
\begin{tabular}{|c|c|c|c|c|c|}
\hline
Ecoregion III & local vs close & close vs far & far vs dissimilar & close vs dissimilar & N \\ \hline
5.3.1 & 1.94E-05 & 3.68E-18 & 5.53E-11 & 4.99E-19 & 135 \\ \hline
8.1.7 & 1.29E-11 & 1.53E-07 & 6.42E-01 & 6.80E-04 & 68 \\ \hline
8.2.3 & 1.33E-03 & 2.46E-02 & 1.01E-01 & 8.52E-04 & 93 \\ \hline
8.2.4 & 1.98E-02 & 6.04E-09 & 1.61E-06 & 9.84E-12 & 67 \\ \hline
8.3.1 & 3.94E-02 & 2.27E-03 & 3.38E-06 & 6.32E-09 & 72 \\ \hline
8.3.4 & 8.39E-09 & 1.54E-03 & 3.47E-01 & 1.61E-04 & 127 \\ \hline
8.3.5 & 9.96E-05 & 1.85E-01 & 3.16E-02 & 5.13E-04 & 97 \\ \hline
8.4.1 & 2.58E-14 & 8.74E-02 & 1.19E-03 & 1.78E-04 & 107 \\ \hline
8.4.2 & 1.72E-03 & 2.64E-01 & 1.69E-02 & 2.37E-05 & 61 \\ \hline
8.5.3 & 1.67E-12 & 8.02E-11 & 6.56E-02 & 1.19E-13 & 90 \\ \hline
9.2.3 & 2.26E-13 & 2.71E-01 & 2.78E-10 & 9.93E-10 & 102 \\ \hline
9.4.2 & 4.70E-04 & 9.32E-01 & 4.81E-01 & 3.24E-01 & 82 \\ \hline
All & 1.70E-63 & 4.19E-32 & 1.34E-24 & 1.54E-66 & 1101 \\ \hline
\end{tabular}
\subcaption*{Panel B: p-value for NSE}
\begin{tabular}{|c|c|c|c|c|c|}
\hline
Ecoregion III & local vs close & close vs far & far vs dissimilar & close vs dissimilar & N \\ \hline
5.3.1 & 2.80E-22 & 1.90E-18 & 1.87E-01 & 8.47E-12 & 135 \\ \hline
8.1.7 & 6.17E-08 & 4.03E-01 & 1.41E-01 & 1.25E-01 & 68 \\ \hline
8.2.3 & 9.86E-01 & 4.13E-02 & 2.67E-04 & 1.77E-04 & 93 \\ \hline
8.2.4 & 6.48E-01 & 1.63E-08 & 2.31E-07 & 2.43E-11 & 67 \\ \hline
8.3.1 & 2.81E-01 & 3.04E-01 & 9.78E-06 & 1.99E-04 & 72 \\ \hline
8.3.4 & 3.37E-13 & 6.56E-01 & 3.41E-03 & 3.18E-02 & 127 \\ \hline
8.3.5 & 6.25E-05 & 3.58E-01 & 7.23E-04 & 1.87E-03 & 97 \\ \hline
8.4.1 & 1.44E-16 & 5.32E-13 & 1.20E-07 & 1.14E-05 & 107 \\ \hline
8.4.2 & 2.30E-03 & 1.77E-05 & 3.72E-04 & 2.98E-03 & 61 \\ \hline
8.5.3 & 5.28E-10 & 1.20E-01 & 4.89E-10 & 3.33E-03 & 90 \\ \hline
9.2.3 & 4.27E-14 & 1.11E-14 & 2.55E-14 & 6.63E-01 & 102 \\ \hline
9.4.2 & 4.77E-03 & 8.08E-01 & 3.30E-01 & 1.32E-01 & 82 \\ \hline
All & 7.35E-69 & 1.85E-01 & 2.23E-36 & 1.27E-14 & 1101 \\ \hline
\end{tabular}
\label{tab_sim_q}
\end{table}

\begin{table}[]
\caption{P-value in wilcoxon signed-rank test for similar vs dissimilar soil moisture experiments when controlling for training size. For space reasons, ``local + close" was abbreviated as ``close'' and similarly for ``local+far'' and ``local+dissimilar''.}
\subcaption*{Panel A: p-value for RMSE}
\begin{tabular}{|c|c|c|c|c|c|}
\hline
Ecoregion III & local vs close & close vs far & far vs dissimilar & close vs dissimilar & N \\ \hline
8.3.5 & 3.55E-09 & 8.44E-07 & 9.35E-09 & 2.10E-11 & 60 \\ \hline
9.3.1 & 5.71E-08 & 2.28E-06 & 3.28E-05 & 1.85E-09 & 52 \\ \hline
9.3.3 & 5.10E-05 & 2.97E-11 & 1.52E-06 & 3.41E-12 & 68 \\ \hline
9.4.1 & 2.32E-09 & 4.72E-09 & 1.55E-08 & 1.63E-10 & 55 \\ \hline
9.4.2 & 3.83E-07 & 8.62E-10 & 1.20E-04 & 1.11E-10 & 55 \\ \hline
10.1.5 & 4.17E-05 & 1.95E-01 & 1.87E-01 & 1.94E-02 & 61 \\ \hline
10.2.4 & 5.65E-05 & 8.75E-01 & 4.06E-11 & 1.93E-11 & 78 \\ \hline
All & 7.05E-41 & 7.08E-34 & 1.36E-35 & 5.38E-58 & 429 \\ \hline
\end{tabular}
\subcaption*{Panel B: p-value for correlation}
\begin{tabular}{|c|c|c|c|c|c|}
\hline
Ecoregion III & local vs close & close vs far & far vs dissimilar & close vs dissimilar & N \\ \hline
8.3.5 & 9.35E-09 & 3.04E-02 & 2.99E-01 & 1.28E-01 & 60 \\ \hline
9.3.1 & 9.00E-08 & 2.63E-01 & 3.92E-01 & 7.99E-01 & 52 \\ \hline
9.3.3 & 2.40E-04 & 2.55E-05 & 1.41E-01 & 2.24E-04 & 68 \\ \hline
9.4.1 & 2.66E-10 & 1.72E-06 & 4.07E-04 & 1.62E-01 & 55 \\ \hline
9.4.2 & 3.07E-07 & 2.66E-10 & 6.17E-02 & 5.64E-10 & 55 \\ \hline
10.1.5 & 1.00E-09 & 4.78E-02 & 9.82E-03 & 2.93E-01 & 61 \\ \hline
10.2.4 & 8.11E-03 & 6.59E-03 & 3.39E-04 & 6.42E-07 & 78 \\ \hline
All & 2.44E-44 & 3.32E-14 & 0.836737 & 4.61E-14 & 429 \\ \hline
\end{tabular}
\label{tab_sim_sm_lim}
\end{table}

\subsubsection{Dataset size controlled, streamflow only}
\label{sec_similar_control}
The previous set of similar vs. dissimilar experiments did not control for the training data size. Hence, one possible explanation for those results is that perhaps data points from dissimilar regions were of lower quality (as compared to data points from similar regions, for the purpose of building a model for a given ROI) but the model was able to overcome this disadvantage due to the vast quantity of dissimilar data available to be added, as compared to the lesser quantity of similar, nearby data available.

\begin{figure}[h]
\includegraphics[width=1\linewidth]{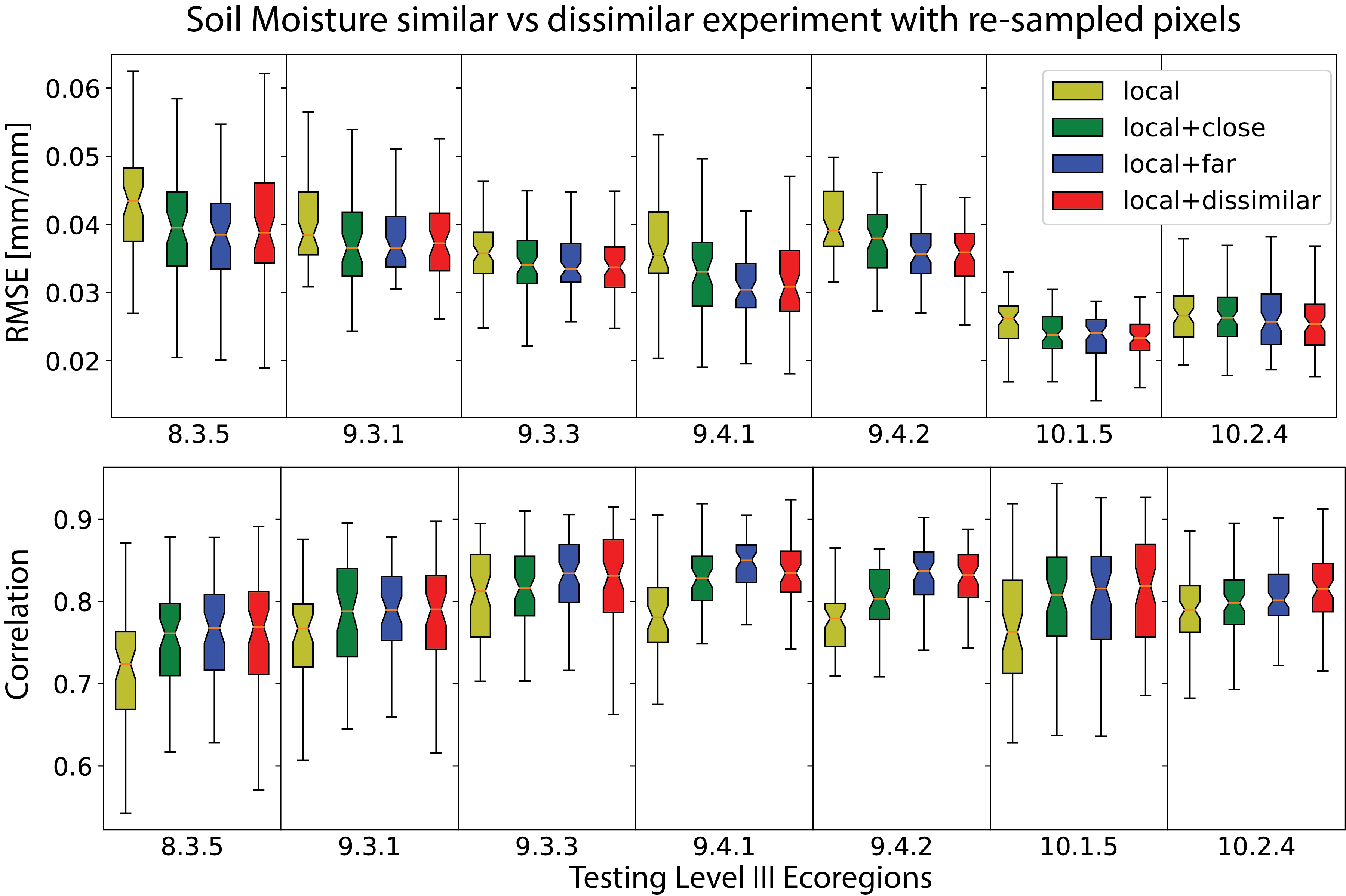}
\caption{Error matrices of soil moisture prediction similar vs dissimilar experiment while controlling data size (Figure \ref{fig_sim_sm} shows the experiments when data size is not controlled). The training regions are re-sampled so that the training set for local+close, local+far and local+dissimilar regions contains same number of pixels.}
\label{fig_sim_lim_sm}
\end{figure}

In the case of soil moisture, where pixels are distributed evenly and continuously across the land surface, it was possible to control for data size while maintaining the characteristics of the close, far, and dissimilar datasets, by sampling (uniformly and at random) equally-sized subsets of each of these datasets (which was not possible for streamflow). Thus for soil moisture prediction, we conducted the same experiments with the resampled datasets, so that this time the ``local+close'', ``local+far'', and ``local+dissimilar'' datasets each were the same size (Figure \ref{fig_sim_lim_sm}). Differences in performance between the alternatives was significantly dampened but still noticeable. Due to the small effect and small per-region sample size most pairwise tests were still significant, but a larger fraction of tests did not identify statistically significant differences (Table \ref{tab_sim_sm_lim}). When all the data were pooled, it is clear that the improvement of ``local+far'' over ``local + close'' was significant, as was the improvement of ``local+dissimilar'' over ``local + close''. However, interestingly, there was no evidence to suggest that there was a meaningful difference between ``local+dissimilar'' vs. ``local+far''.

These results allow us to reject the notion that the ``far'' and ``dissimilar'' data points were low quality for building a model at  given ROI. Combined with the uncontrolled data experiments, we see that diversity in data helps due to both quantity and characteristics of these data points (with the former showing a larger effect). More diverse data is clearly better than more homogeneous data, further supporting (but not proving) the hypothesis that heterogeneity in the data has a regularizing effect that could reduce overfitting.

%
Overall, the experiments together show an inherent benefit of allowing more heterogeneous inputs to deep learning models in hydrology -- not only do the heterogeneous inputs appear to help the model, but hetergeoenous datasets are also much more plentiful and provide
 opportunities to amass much larger datasets. This observation liberates us from using small, regionalized datasets when practicing deep learning in hydrology and (in our opinion) should not be understated. 

Another possible way to explain the advantage is more dissimilar datasets expanded the range of inputs during training. Or, they covered more parts of the input space, so future inferences became more of an interpolation than extrapolation. It is well-known in hydrology that we can ``trade space for time", e.g., we can examine the experience of a catchment's southern, warmer neighbor to project its future trajectory after global warming. The fact that DL can assimilate all the data together perhaps makes it more robust for future projections.

\section{Further Discussion}
\label{sec_further}
The global vs. local and similar vs. dissimilar experiments showed that, when using state-of-the-art prediction tools, for soil moisture and stream flow prediction it is better to pool data together rather than to stratify and build separate models. There are several (not mutually exclusive) explanations for this effect. One is that heterogeneous data may provide a regularizing effect that reduces overfitting. Another is that a deep learning model may use its internal representation to construct a multi-level model that captures similarities among regions (e.g., the main effect) as well as region-specific differences.
%
In the latter case, it would suggest that deep networks may have extracted the common part of the data and built a basic soil moisture dynamics model, e.g., it may know that soil moisture rises when rainfall occurs and declines when rainfall ceases. It could also have specialized into the different response curves as modulated by different soil and land use characteristics. When data came from more different regions, it was easier for the model to discern the most basic, most fundamental responses for the basic-level component. When data came from more similar regions, there is more commonality, but such commonality may not be fundamental to the problem. 

The data synergy effects seemed less pronounced with streamflow predictions. An explanation is there are are more latent processes with rainfall-runoff modeling, e.g., we do not have accurate-enough inputs about geology, e.g., aquifer laying and transmissivity, and the representation of stream network and routing process were greatly simplified. Due to these unknown and potentially confounding factors, it would be more difficult for the network to extract the true multilevel model. This situation would not be unique to streamflow prediction, and may also apply to stream temperature modeling \cite{Rahmani2020}, water chemistry \cite{Zhi2020}, etc. It could be said that most geoscientific variables, to some extent, have latent variable or parameters that we cannot fully describe. Hence, we caution against generalizing data synergy in the absolute sense to all stratification schemes and to all problems. However, pooling big data together is certainly one option that can be tried to improve performance.

We also note that we have not  ``proved'' the multilevel theory or the regularization theory although they both support the observed experimental results. Additional study of the network parameters themselves would be needed to confirm either theory.

On a side note, DL model complexity is not equal to the number of parameters in the network. For instance, Piantadosi (2018) \cite{Piantadosi2018} showed that a 1-parameter function can fit any 2-dimensional dataset to any desired level of precision (i.e., a one-parameter function can overfit a 2-dimensional dataset of any size). Additionally, recent research showed the existence of the ``double descent curve" for deep learning models \cite{Belkin2019} In this curve, empirical observations show that in the under-parameterized regime, test error first decreases, then starts to increase as we approach the inflection point. However as the number of parameters keeps increasing, the error undergoes a second descent. While the exact reasons for this behavior are still being investigated, there is evidence that alternative methods for controlling complexity (aside from the number of parameters) plays a key role. Such methods are known as regularization and include adding penalties in the objective function for large weights, dropout \cite{Srivastava2014}, and early stopping (using a holdout dataset to determine when to stop training). In general, such methods make a model smoother (small changes to the input do not tend to cause large changes in the output). In fact, non-smoothness is the reason why a 1 parameter function can overfit any 2-dimensional dataset \cite{Piantadosi2018}.

\section{Conclusion}

In this paper, we introduced and studied the data synergy effects in predicting soil moisture and streamflow using LSTM networks, which can be concisely described as: more data and diversity are helpful in improving model performance. On the practical level, the data synergy effects guides us in dataset construction and processing: unless we fundamentally lack critical inputs, we should not assume stratification is the best approach; rather, we should try to compile a large dataset, from diverse domains, and attempt a unified model. If data collection budget is limited, we can first collect enough local data to build a robust model with reasonable performance, and then collection data from heterogeneous sources. While it cannot be guaranteed that the performance will be better -- for problems where the DL model itself performs very poorly or there are critical variables that are not known, stratification may nonetheless be useful -- our experiences suggest the chances are high that a more diverse dataset leads to a more robust and more accurate model. On a philosophical level, if we only have a small dataset and have to build a machine learning model specifically for this model. It should not be expected that the model provides an optimal prediction or captures a universal relationship. In case of truly heterogeneous inputs that are not comparable, other approaches such as transfer learning is applicable \cite{Ma2020}.

Among all the experiments we tried, there are no cases found that the model would performs worse after training with regions outside the ROI. DL models would not be confused by additional information, even when they appears to be unrelated by our judgement. In fact, as the similar vs dissimilar experiment shows, dissimilar ecoregions may bring in more knowledge compared to the similar ones. 
The exact mechanism by which DL models accomplish this is not yet known, but we hypothesis that it  may be related to multilevel models.
We also note that allowing more heterogeneous data, by default, admits a much greater amount of data, which could be an important reason why big data machine learning techniques improve performance. Hence both of these effects (quality due to heterogeneity and quantity due to heterogeneity) are important.

\bibliographystyle{unsrt}  
\bibliography{main}
\end{document}